%% file: root.tex
\documentclass[letterpaper, 10 pt, conference]{ieeeconf}  

\IEEEoverridecommandlockouts                              

\usepackage{cite}
\usepackage{graphicx}
\usepackage{xspace}
\usepackage{amsmath}
\usepackage{amssymb}
\usepackage{multirow}
\usepackage{booktabs}
\usepackage{colortbl}
\usepackage{xcolor}
\definecolor{cacrowgray}{RGB}{235,235,235}
\usepackage{subcaption}

\newcommand{\sysname}{\textit{CAC-VLA}\xspace}

\title{\LARGE \bf
CAC-VLA: Context-Gated Action Conditioning for Vision-Language-Action Models
}

\author{
Yifu Xiong\textsuperscript{1*}, Wenhao Yu\textsuperscript{1*}$^\dagger$, Jiaxuan Lin\textsuperscript{1}, Bojun Zou\textsuperscript{1}, Jiahao Li\textsuperscript{1 } \\
   Lu Zhang\textsuperscript{2}$^\ddagger$, Yanyong Zhang\textsuperscript{1}, Jianmin Ji\textsuperscript{1}$^\ddagger$ \\
  \textsuperscript{1}University of Science and Technology of China (USTC) \\
  \textsuperscript{2}{Institute of Artificial Intelligence, Hefei Comprehensive National Science Center}
\thanks{* Equal contributions, $\dagger$ Project leader, $\ddagger$ Corresponding authors}
\thanks{$^{1}$University of Science and Technology of China (USTC)}%
\thanks{$^{2}$Institute of Artificial Intelligence, Hefei Comprehensive National Science Center}%
}

\begin{document}

\maketitle
\thispagestyle{empty}
\pagestyle{empty}

\begin{abstract}

Vision-Language-Action (VLA) models have become a promising paradigm for generalist robot manipulation, where visual-language representations are used to condition continuous action generation. However, these representations are not explicitly optimized for action conditioning, leaving the action expert to bridge the gap between multimodal understanding and precise motor control. Recent action-reasoning methods introduce additional modules to generate explicit action plans or action-space reasoning signals, demonstrating the benefit of action-level guidance but often requiring separate action-generation frameworks. We propose \sysname, a Context-Gated Action Conditioning framework that learns a lightweight latent-action interface directly within the VLM. Instead of generating executable trajectories, \sysname trains the VLM to predict coarse-to-fine latent actions, which are structured representations encoded from future action segments, and adaptively leverages them to condition the action expert via a context gate. This enables VLM-native action conditioning while calibrating the influence of latent-action guidance on expert action generation. Experiments on LIBERO, LIBERO-Plus, and CALVIN demonstrate the effectiveness of \sysname, achieving 98.9\% and 90.4\% average success rates on LIBERO and LIBERO-Plus, respectively, and outperforming \(\pi_{0.5}\) by 9.3 percentage points on CALVIN.

\end{abstract}

\input{figure/overview}

\section{INTRODUCTION}

Vision-Language-Action (VLA) models have emerged as a promising paradigm for generalist robot manipulation. A VLA policy receives visual observations and language instructions, and generates executable actions for interacting with the physical world. By combining pretrained Vision-Language Models (VLMs) with action decoders or action experts, recent VLA models have demonstrated the ability to follow diverse instructions and generalize across objects, scenes, and manipulation tasks~\cite{paligemma, swiftVLA, dreamvla, pi05, pi0}.

Despite this progress, the interface between multimodal understanding and continuous control remains largely implicit in many existing VLA architectures. The VLM produces representations that capture visual and linguistic information, while the action expert is responsible for transforming these representations into low-level robot commands. However, the representations passed from the VLM to the action expert are not explicitly organized around future action structure. As a result, the action expert must infer both the task-level motion intent and the fine-grained control signal from the same visual-language context. This implicit transformation can be particularly challenging for long-horizon manipulation, where the appropriate action depends not only on the current observation but also on the temporal structure of the actions that should follow.

A natural way to address this issue is to introduce an intermediate signal that provides action-related information before continuous action generation. Language-based approaches use textual subgoals or task plans to decompose manipulation tasks~\cite{pi05, saycan}, while vision-based approaches predict goal images or future observations to guide the policy~\cite{cot_vla, worldvla, dreamvla}. More recent methods move the intermediate representation closer to the control space by generating action plans, reference trajectories, or action-space priors~\cite{acot_vla}. These methods suggest that action-structured guidance can reduce the burden on the action decoder or expert. However, these approaches typically obtain intermediate guidance through auxiliary reasoning or action-generation pathways, leaving open how its influence should be adjusted as the action expert refines the current action chunk. In an iterative action-generation process, the relevance of action guidance may change across refinement steps, while inaccurate guidance can steer the predicted action toward an undesired trajectory if injected too strongly. This motivates a mechanism that can adaptively regulate the contribution of action guidance during generation.

We propose \sysname, a Context-Gated Action Conditioning framework that introduces an explicit latent-action interface within the VLA policy. Given a visual observation and a language instruction, dedicated query tokens in the VLM predict latent representations of future action segments. These latent actions are not directly executed; instead, they are provided to the action expert as an intermediate representation of future action structure.

To construct the training targets, we encode future robot action segments into ordered latent representations~\cite{oat}. The VLM is trained to predict these representations from the current visual-language context. At inference time, future action segments are unavailable, and the expert receives only the latent actions predicted by the VLM. This design allows the VLM to provide action-structured information without introducing an additional executable action-generation branch. Since its relevance may vary across flow steps and manipulation phases, \sysname uses cross-attention to retrieve latent-action information relevant to the current action state and a context gate to modulate the resulting residual update before it is injected into the action expert. This adaptive injection allows the expert to exploit latent-action guidance while preserving its capacity for continuous action modeling.

We evaluate \sysname on the LIBERO, LIBERO-Plus, and CALVIN D-D manipulation benchmarks. \sysname achieves an average success rate of 98.9\% on LIBERO and 90.4\% on the supervised fine-tuning setting of LIBERO-Plus. On CALVIN D-D, \sysname improves the 5-task completion rate from 57.8\% to 67.1\% and increases the average sequence length from 3.639 to 3.944 compared with $\pi_{0.5}$. We further conduct ablation studies on the latent-action horizon and context-gated conditioning, and evaluate the method on real-world pick-and-place and block-stacking tasks. These experiments provide evidence for the usefulness of the proposed latent-action interface under the evaluated simulation and real-world settings.

Our contributions are summarized as follows:
\begin{itemize}
    \item We introduce a latent-action interface for explicitly conveying future action structure from the VLM to a continuous action expert without generating executable intermediate trajectories.
    \item We propose \sysname, a context-gated action conditioning framework that adaptively modulates the influence of VLM-predicted latent actions during continuous action generation.
    \item We evaluate the proposed framework on LIBERO, LIBERO-Plus, CALVIN D-D, and real-world manipulation tasks, with ablations that analyze the effects of latent-action horizon and context-gated conditioning.
\end{itemize}

\section{Related Work}

\subsection{Vision-Language-Action Models}

Vision-Language-Action models combine visual-language understanding with robot action generation. Early approaches such as PaLM-E~\cite{palm_e} connect large-scale multimodal representations with embodied control. More recent systems adapt pretrained vision-language backbones and pair them with action decoders, action heads, or action experts to support generalist manipulation~\cite{openvla, octo, pi0, pi05, yu2024ldp, duan2025stdarm, gao2026drift}. These models mainly rely on the visual-language representation as the interface for downstream action prediction. \sysname follows this general VLA paradigm but introduces an explicit latent-action interface between the VLM and the continuous action expert.

\subsection{Action-Level Guidance}

Several recent works introduce intermediate signals that are closer to the action space than language or visual representations. ACoT-VLA~\cite{acot_vla} explores action-level reasoning through coarse action intents, reference trajectories, and latent action priors. Coarse-to-Control~\cite{coarse_to_control} adopts a plan-and-execute formulation in which coarse action tokens are generated before executable action tokens, with planning and execution represented in a shared action-token space. Flowing With Purpose~\cite{lafm} uses predicted latent motion primitives to select structured source distributions for a flow-matching policy. These methods demonstrate different ways of using action-related intermediate representations to support robot control. In contrast, \sysname predicts latent actions from dedicated VLM query tokens and uses a context gate to adaptively inject them into a continuous action expert, without treating the intermediate representation as an executable action sequence.

\subsection{Latent Actions and Action Tokenization}

Latent actions and action tokenization have been studied as compact representations for modeling robot behavior. LAPA~\cite{lapa} learns discrete latent actions from unlabeled videos and uses them for VLA pretraining. UniVLA~\cite{univla} extracts task-centric latent actions from heterogeneous videos to support cross-embodiment policy learning. Other work investigates discrete action tokenization for representing robot actions with compact sequences~\cite{pertsch2025fast}. Ordered Action Tokenization (OAT)~\cite{oat} further provides an ordered latent representation for action segments.

\sysname differs from these approaches in both the source and the role of the latent representation. Rather than using latent actions primarily for video-based pretraining, cross-embodiment transfer, or direct action decoding, \sysname uses OAT-encoded future robot action segments as supervision for VLM query tokens. The predicted latent actions are not treated as an executable action space. Instead, they serve as non-executable conditioning signals for a continuous action expert, whose use of the latent guidance is modulated by the proposed context gate.

\section{Method}

We present \sysname, a framework that equips VLA models with a VLM-native latent-action interface and context-gated expert conditioning. We first formulate latent-action conditioned policy learning in Sec.~\ref{sec:problem}, then describe latent action prediction in Sec.~\ref{sec:latent_prediction}, context-gated action conditioning in Sec.~\ref{sec:cgac}, and the training and inference procedure in Sec.~\ref{sec:train_infer}.

\subsection{Problem Formulation}
\label{sec:problem}
Given a visual observation \(o_t\) and a language instruction \(l\), a VLA policy aims to predict an executable action chunk
\(\mathbf{a}_{t:t+H_e-1}\), where \(H_e\) denotes the action horizon. A standard VLA model directly maps the visual-language input to actions:
\begin{equation}
    \mathbf{a}_{t:t+H_e-1}
    =
    \pi_{\theta}
    \left(
    o_t, l
    \right).
\end{equation}

In this work, we introduce a latent-action representation \(\mathbf{z}_t\) as an intermediate action-conditioned interface between the VLM and the action expert. Instead of relying only on visual-language representations, the policy first predicts latent actions from the multimodal context and then conditions the action expert on them:
\begin{equation}
    \mathbf{z}_t
    =
    f_{\theta}
    \left(
    o_t, l
    \right),
    \qquad
    \mathbf{a}_{t:t+H_e-1}
    =
    \pi_{\theta}
    \left(
    o_t, l, \mathbf{z}_t
    \right).
\end{equation}
Here, \(f_{\theta}\) denotes the VLM-side latent-action predictor, and \(\mathbf{z}_t\) provides action-structured conditioning for continuous expert control.

\subsection{VLM-native Latent Action Prediction}
\label{sec:latent_prediction}

To provide the action expert with structured action information, we train dedicated VLM query tokens to predict latent actions. This design allows the VLM to learn an action-conditioned representation from visual-language context, while leaving continuous control to the action expert.

\paragraph{Choice of action tokenizer.}
The choice of the action tokenizer is important because it determines the representation predicted by the VLM and the information subsequently provided to the action expert. For our latent-action conditioning interface, the representation should be compact and fixed-structured, while preserving an ordered organization from coarse motion structure to finer action details. These properties are preferable to directly using variable-length discrete action sequences~\cite{pertsch2025fast}, which introduce additional sequence modeling and detokenization considerations. We therefore use OAT, whose ordered latent representation provides a natural coarse-to-fine target for VLM prediction and can be directly projected into the action expert's conditioning space.

To construct supervision for these latent actions, OAT encodes future action segments into latent representations. Given a future action segment $\mathbf{a}_{t:t+H_l-1}$, the tokenizer produces raw latent vectors:
\begin{equation}
    \mathbf{z}^{\mathrm{oat}}_t
    =
    \mathrm{OAT}
    \left(
    \mathbf{a}_{t:t+H_l-1}
    \right),
\end{equation}
where $H_l$ denotes the latent-action horizon. Unlike the expert horizon $H_e$, $H_l$ can be flexibly configured, allowing the latent representation to summarize action structure over different temporal ranges and provide broader temporal guidance for generating the current action chunk.

We append learnable latent query tokens $\mathbf{q}$ to the VLM input to elicit latent-action predictions from the visual-language context. As shown in Fig.~\ref{fig:overview}(b), we use an expert-aware attention mask that prevents action expert tokens from directly attending to these query tokens. This ensures that the query tokens affect the expert only through the predicted latent actions and the subsequent context-gated conditioning module.

After processing the visual-language context and the latent queries, the VLM produces query hidden states:
\begin{equation}
    \mathbf{h}^{q}_t
    =
    \mathrm{VLM}
    \left(
    o_t, l, \mathbf{q}
    \right).
\end{equation}
These query states are projected to the raw OAT latent space through a lightweight prediction head:
\begin{equation}
    \hat{\mathbf{z}}_t
    =
    W_z
    \mathrm{LN}
    \left(
    \mathbf{h}^{q}_t
    \right).
\end{equation}

We supervise the predicted latent actions with a token-wise Smooth-L1 loss over the raw OAT latent dimension:
\begin{equation}
    \mathcal{L}_{\mathrm{align}}
    =
    \frac{1}{\sum_i m_i}
    \sum_i
    m_i
    \cdot
    \operatorname{SmoothL1}
    \left(
    \hat{\mathbf{z}}_{t,i},
    \mathbf{z}^{\mathrm{oat}}_{t,i}
    \right),
    \label{eq:align}
\end{equation}
where $m_i$ denotes the mask of the $i$-th latent token.

\subsection{Context-Gated Action Conditioning}
\label{sec:cgac}

Although latent actions encode structured future-action information, they should be used as guidance rather than fixed commands. A simple strategy is to add or concatenate latent-action features with action hidden states, but such fixed fusion applies the same conditioning strength regardless of the expert layer or the current action-generation state. Since the action expert progressively refines continuous action representations, the usefulness of latent-action information may vary during generation.

We therefore introduce \emph{Context-Gated Action Conditioning}. The module first projects latent actions into expert-compatible conditioning tokens and uses cross-attention to retrieve latent-action information conditioned on the current action hidden states. It then applies a context gate to control how much of the retrieved update is injected into the expert through a residual connection.

The conditioning source differs between training and inference. During training, we use the encoded OAT latent target to provide stable action-structured conditioning; during inference, this source is replaced by the VLM-predicted latent action. We define

\begin{equation}
    \mathbf{z}^{c}_t =
    \begin{cases}
    \mathbf{z}^{\mathrm{oat}}_t, & \mathrm{training},\\
    \hat{\mathbf{z}}_t, & \mathrm{inference},
    \end{cases}
    \qquad
    \mathbf{c}_t
    =
    \phi_c
    \left(
    \mathbf{z}^{c}_t
    \right),
\end{equation}
where \(\phi_c(\cdot)\) is a token-wise linear projection into the action expert hidden space. The resulting \(\mathbf{c}_t\) serves as latent-action conditioning tokens for expert cross-attention.

We apply the conditioning module after self-attention and before the feed-forward block in each action expert layer. Let \(\mathbf{x}^{l}\in\mathbb{R}^{B \times H_e \times D}\) be the action hidden states at layer \(l\). We first normalize them as
\begin{equation}
    \bar{\mathbf{x}}^{l}
    =
    \mathrm{Norm}
    \left(
    \mathbf{x}^{l}
    \right).
\end{equation}
The normalized action states query the latent-action conditioning tokens through cross-attention:
\begin{equation}
    \mathbf{u}^{l}
    =
    \mathrm{CrossAttn}
    \left(
    Q=\bar{\mathbf{x}}^{l},
    K=\mathbf{c}_t,
    V=\mathbf{c}_t
    \right)
\end{equation}
where \(\mathbf{u}^{l}\) denotes the latent-action update retrieved by the expert.

To determine how strongly this update should affect the expert, we compute a context gate conditioned on the current action-generation context, represented by the normalized action state.
\begin{equation}
    \mathbf{g}^{l}
    =
    \sigma
    \left(
    W_g
    \tanh
    \left(
    W_x \mathbf{s}^{l}_{x}
    \right)
    \right),
    \mathbf{s}^{l}_{x}
    =
    \mathrm{AvgPool}
    \left(
    \bar{\mathbf{x}}^{l}
    \right).
    \qquad
\end{equation}
where \(\mathbf{g}^{l}\in\mathbb{R}^{B \times 1 \times D}\) is a channel-wise gate shared across all action tokens. Finally, the retrieved latent-action update is injected through a gated residual connection:
\begin{equation}
    \mathbf{x}^{l}_{+}
    =
    \mathbf{x}^{l}
    +
    \mathbf{g}^{l}
    \odot
    \mathbf{u}^{l}.
\end{equation}

This design separates retrieval from injection. Cross-attention retrieves latent-action information conditioned on the current action state, while the context gate determines how much of the retrieved update should be added back to the expert representation. Compared with fixed fusion, this adaptive residual injection better preserves the expert's continuous action modeling capacity.

\subsection{Training and Inference}
\label{sec:train_infer}

The training objective combines the original action expert loss with the latent alignment loss:
\begin{equation}
\mathcal{L} = \mathcal{L}_{\mathrm{act}} + \lambda_{\mathrm{align}}\mathcal{L}_{\mathrm{align}}
\end{equation}
where $\mathcal{L}_{\mathrm{act}}$ denotes the original flow-matching action expert loss and $\lambda_{\mathrm{align}}$ balances latent-action supervision. During training, the frozen OAT tokenizer encodes future action segments as both alignment targets and expert conditioning sources. The VLM query tokens are trained to predict the same latent representation through $\mathcal{L}_{\mathrm{align}}$. During inference, the OAT tokenizer is removed, and the expert is conditioned only on VLM-predicted latent actions. Thus, \sysname uses future-action structure as training-time supervision while remaining VLM-native at deployment.

\input{libero}
\input{libero_plus}
\input{calvin}

\section{Experiments}
\label{sec:exp}

We evaluate \sysname on standard simulation benchmarks to validate the effectiveness of VLM-predicted latent actions for continuous robot control. Our experiments compare \sysname with existing VLA methods that rely on visual, linguistic, or action-level guidance, and further analyze the contribution of the proposed latent-action conditioning design through ablations on latent-action horizon and context gating.

\subsection{Experimental Setup}
\label{sec:exp_setup}

\paragraph{Benchmarks and metrics.}
We evaluate \sysname on the LIBERO~\cite{libero}, LIBERO-Plus~\cite{libero_plus}, and CALVIN D-D~\cite{calvin} manipulation benchmarks. LIBERO contains four multi-task manipulation suites, including Spatial, Object, Goal, and Long, while LIBERO-Plus evaluates robustness under distribution shifts in camera viewpoint, robot embodiment, language instruction, lighting, background, observation noise, and scene layout. CALVIN D-D evaluates long-horizon language-conditioned manipulation through sequences of consecutive tasks. Following the official evaluation protocols, we report success rate as the main metric on LIBERO and LIBERO-Plus, including per-suite results and the average score, while reporting the completion rates for 1--5 consecutive tasks and the average sequence length on CALVIN D-D.

\paragraph{Implementation.}
We implement \sysname by extending the $\pi_{0.5}$ architecture with VLM-side latent query tokens, an OAT-aligned latent-action prediction head, and context-gated expert conditioning. We use 8 learnable latent query tokens for latent-action prediction. Unless otherwise specified, we set the latent-action loss weight to $\lambda_{\mathrm{align}}=0.1$ and use $H_l=H_e=10$, where $H_l$ and $H_e$ denote the latent-action and executable-action horizons, respectively. For LIBERO, we train the model with a batch size of 128, a learning rate of $2.5\times10^{-5}$, and a total of $4\times10^{4}$ training steps. For LIBERO-Plus, we use a batch size of 128, a learning rate of $1.25\times10^{-5}$, and train for $3\times10^{4}$ steps. For CALVIN D-D, we use a batch size of 32, a learning rate of $1.25\times10^{-5}$, and train for $3\times10^{4}$ steps. The OAT encoder is trained from scratch for $2\times10^{3}$ steps using the same training configuration as the official implementation. Unless otherwise specified, we follow the corresponding benchmark protocols for data preprocessing and evaluation.

\subsection{Simulation Experiments}
\label{sec:sim_exp}

We next report our method on LIBERO~\cite{libero}, LIBERO-Plus~\cite{libero_plus}, and the CALVIN D-D~\cite{calvin} split.

\paragraph{Results on LIBERO.}
Table~\ref{tab:libero} reports the comparison on the standard LIBERO benchmark. \sysname achieves the highest average success rate of 98.9\%. These results indicate that the proposed latent-action interface effectively supports continuous expert control across diverse standard manipulation tasks.

\paragraph{Results on LIBERO-Plus.}
Table~\ref{tab:libero_plus} reports the results on LIBERO-Plus under both zero-shot transfer and supervised fine-tuning. Under supervised fine-tuning, \sysname achieves the highest average success rate of 90.4\%. These results demonstrate the robustness of \sysname under distribution shifts.

\paragraph{Results on CALVIN D-D.}
Table~\ref{tab:calvin} reports the results on the CALVIN D-D benchmark. Compared with $\pi_{0.5}$, \sysname reports higher values on all reported CALVIN metrics. In particular, it improves the 5-task completion rate from 57.8\% to 67.1\% and increases the average sequence length from 3.639 to 3.944, demonstrating stronger performance on long-horizon manipulation sequences.

\input{abl_horizon}
\input{abl_module}

\subsection{Ablation Studies}
\label{sec:ablation}

We conduct ablation studies to analyze two key design choices in \sysname: the latent-action horizon and the context-gated conditioning mechanism.

\paragraph{Effect of latent-action horizon.}
As shown in Table~\ref{tab:ablation_libero_plus_horizon}, $H_l=10$ achieves the highest average success rate of 90.43\% in the current LIBERO-Plus setting, compared with 88.38\% for $H_l=20$ and 87.66\% for $H_l=30$. These results suggest that increasing the latent-action horizon does not necessarily improve performance under this training and evaluation configuration, and that excessively long horizons may introduce less relevant or noisier future information for the current action chunk. 

\paragraph{Effect of context gating.}
Table~\ref{tab:ablation_libero_plus_module} compares the full model with variants that remove context gating or latent-action conditioning. Although the absolute improvement from context gating is relatively modest, LIBERO-Plus contains 10,030 task instances spanning seven perturbation dimensions and 21 sub-dimensions~\cite{libero_plus}. Thus, the reported average scores aggregate performance over a broad and systematically constructed evaluation set rather than a small number of trials. Removing the context gate reduces the average success rate from 90.43\% to 89.36\%, corresponding to a decrease of 1.07 percentage points. The ``w/o Latent'' variant still predicts latent actions and retains the latent alignment loss, but does not use the predicted latents to condition the action expert. Its average success rate decreases to 87.95\%, a reduction of 2.48 percentage points from the full model. The overall ordering of the three variants provides evidence that both adaptive gating and latent-action conditioning contribute to the evaluated performance, although formal statistical significance would require multi-seed evaluation or confidence-interval analysis.

\subsection{Analysis of Adaptive Latent-Action Conditioning}
\label{sec:guidance_analysis}

We further analyze \sysname from two perspectives: whether the VLM learns the coarse-to-fine latent action structure, and whether the context gate adaptively modulates the contribution of latent-action conditioning during execution. For the latter analysis, we use the effective latent-conditioning strength as a measure of how strongly the gated latent-action signal contributes to the action expert.

\begin{figure}[t]
    \centering
    \includegraphics[width=\columnwidth]{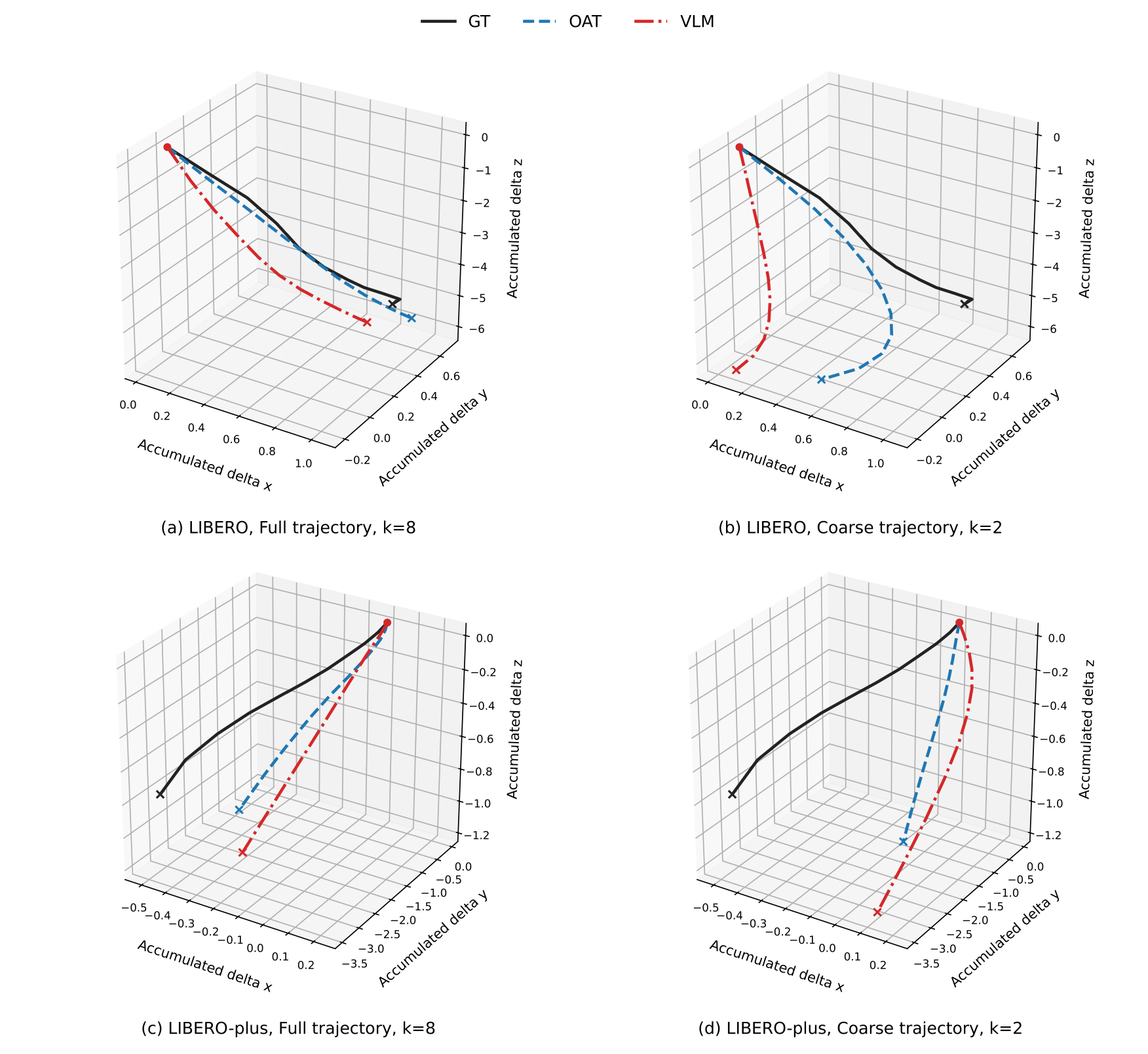}
    \caption{Visualization of coarse-to-fine latent-action prediction using the same model trained on LIBERO. Panels (a) and (b) show the results on LIBERO, while panels (c) and (d) show the zero-shot results on LIBERO-Plus. The full and coarse trajectories correspond to $k=8$ and $k=2$, respectively. ``GT'' denotes the ground-truth action trajectories from the dataset, ``OAT'' denotes the trajectories obtained by encoding the ground-truth action chunks with the OAT encoder and decoding them with the OAT decoder, and ``VLM'' denotes the trajectories obtained by decoding the latent actions predicted by the VLM with the OAT decoder.}
    \label{fig:coarse_to_fine}
    \vspace{-4.0mm}
\end{figure}

\paragraph{Coarse-to-fine latent-action structure.}
Fig.~\ref{fig:coarse_to_fine} compares the ground-truth action trajectories with the trajectories reconstructed from the OAT encoder--decoder and the trajectories obtained by decoding the VLM-predicted latent actions with the OAT decoder. The OAT reconstruction provides a reference for how the encoder--decoder preserves the structure of the ground-truth actions, while the VLM-decoded trajectory indicates whether the VLM predicts latent actions with a similar coarse-to-fine organization. The VLM-decoded trajectories exhibit a similar organization to the OAT reconstructions on both LIBERO and LIBERO-Plus. This structure is also preserved on LIBERO-Plus without additional fine-tuning, even though the test distribution differs from the LIBERO training distribution. These results provide qualitative evidence that the VLM learns a transferable representation of future action chunks rather than merely memorizing patterns specific to the training environment. Although the predicted latent actions are not directly executable, they provide the action expert with an informative estimate of the future action structure.

\begin{figure}[t]
    \centering
    \includegraphics[width=\columnwidth]{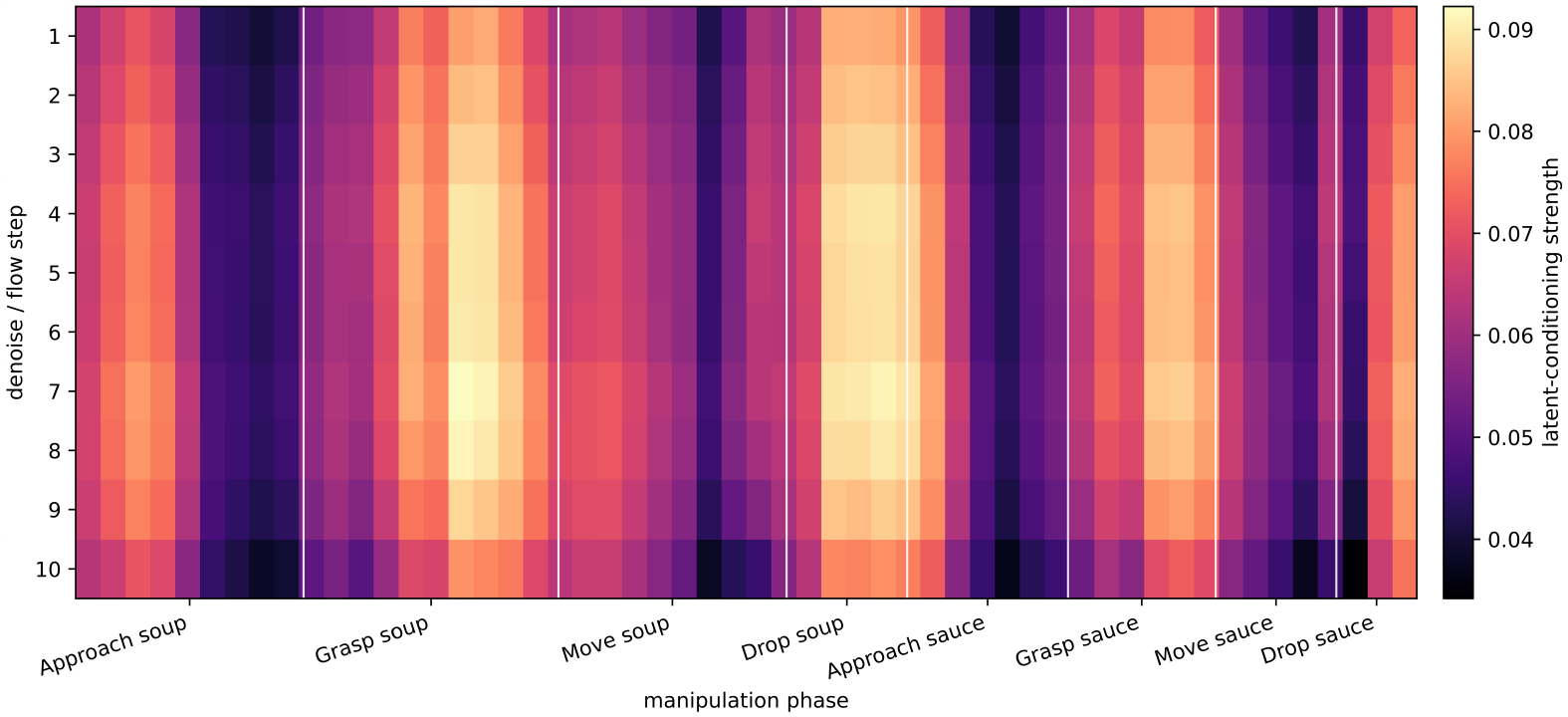}
    \caption{Phase-level visualization of the effective latent-conditioning strength across flow steps and manipulation phases for soup and sauce tasks. The manipulation phases include approach, grasp, move, and drop.}
    \label{fig:latent_conditioning_heatmap}
\end{figure}

\begin{figure}[t]
    \centering
    \includegraphics[width=\columnwidth]{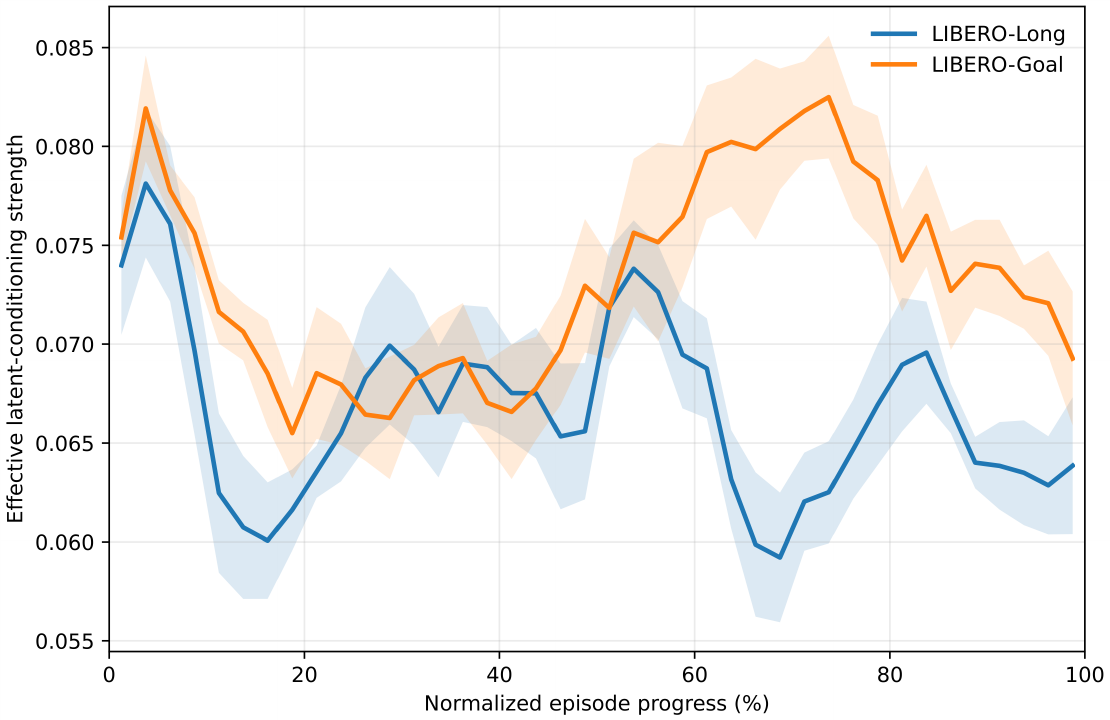}
    \caption{Effective latent-conditioning strength over normalized episode progress on LIBERO-Long and LIBERO-Goal.}
    \label{fig:gate_curve}
    \vspace{-4.0mm}
\end{figure}

\input{figure/real_merge}

\paragraph{Phase-dependent latent-action guidance.}
Figure~\ref{fig:latent_conditioning_heatmap} visualizes the effective latent-conditioning strength across flow steps and manipulation phases. A clear phase-dependent pattern emerges: latent-action guidance is generally stronger during free-space moving phases and becomes weaker during contact-intensive interaction phases. This behavior is consistent with the different control requirements of the two stages. During motion, the future-action latent provides a reliable trajectory prior that captures the global spatial intent of the task, and stronger conditioning helps preserve this intended direction of motion. In contrast, once physical interaction begins, successful execution depends more strongly on the instantaneous robot–object configuration, contact state, and local feedback. Excessively enforcing the nominal future-action prior in these phases could restrict the policy's ability to adapt to contact uncertainty and execution deviations. The reduced conditioning strength therefore suggests that the policy learns to treat latent actions as an adaptive motion prior rather than a rigid command: it follows the planned motion more strongly in free space while allowing greater feedback-driven adaptation during interaction. The variation across flow steps further indicates that this trade-off is dynamically adjusted throughout the action-refinement process.

Fig.~\ref{fig:gate_curve} further shows that the effective latent-conditioning strength changes with normalized episode progress and follows different trends on LIBERO-Long and LIBERO-Goal. This suggests that the context gate adapts latent-action guidance to the progress of execution and the characteristics of the manipulation task. Together, these visualizations suggest that \sysname combines a structured representation of future action chunks with adaptive guidance modulation. The VLM provides an informative action-level signal, while the context gate controls its influence on continuous action generation across different flow steps, task phases, and task types.

\subsection{Real-World Experiments}
\label{sec:real_world}

We further evaluate \sysname on two real-world tabletop manipulation tasks using a UR7e robotic arm: pick-and-place and block stacking. Both tasks use two Intel RealSense D435i RGB-D cameras, including a wrist-mounted camera and a third-person camera, together with robot proprioceptive states and language instructions. For each task, we collect 50 demonstrations and fine-tune \sysname and the $\pi_{0.5}$ baseline separately using the same demonstrations, input modalities, and task-specific training settings. Each method is evaluated for 25 independent trials per task under the same robot, sensing, initialization, and language conditions.

\paragraph{Metrics.}
For pick-and-place, we report Task Score and Full Success Rate (Full SR). A trial receives a Task Score of 1.0 if the robot successfully completes the entire task, 0.5 if it successfully grasps the block but fails to place it inside the target basket, and 0 otherwise. Full SR is the percentage of trials in which the block is successfully placed and remains stably inside the target basket. For block stacking, we report Stacking Success Rate (Stacking SR), which is the percentage of trials in which the manipulated block is released and remains stably supported on top of the target block.

\paragraph{Results.}
As shown in Fig.~\ref{fig:real_results}, \sysname achieves a Task Score of 72\% and a Full SR of 64\% ($16/25$ successful trials) on pick-and-place, compared with 48\% and 16\% ($4/25$ successful trials) for the $\pi_{0.5}$ baseline. On block stacking, \sysname achieves a Stacking SR of 52\% ($13/25$ successful trials), compared with 36\% ($9/25$ successful trials) for $\pi_{0.5}$.Video demonstrations of the real-world pick-and-place and block-stacking experiments are provided in the supplementary video.

Building on the coarse-to-fine analysis above, Fig.~\ref{fig:coarse_to_fine} suggests that the learned latent-action structure remains observable under the evaluated LIBERO-to-LIBERO-Plus distribution shift, even without additional fine-tuning. This observation is relevant to the real-world setting, where the training demonstrations and evaluation configurations are not exactly identical. By providing a coarse but informative representation of the future action chunk, the latent-action interface may help the action expert maintain a more consistent action-generation strategy under changes in visual observations and physical configurations. Taken together, these observations suggest that context-gated latent-action conditioning can provide transferable action-level guidance beyond the training distribution.

\section{CONCLUSIONS}

We presented \sysname, a context-gated action conditioning framework that equips VLA models with a VLM-native latent-action interface for structured guidance during continuous action generation. Experiments show that \sysname achieves 98.9\% and 90.4\% success rates on LIBERO and LIBERO-Plus, respectively, and improves the 5-task completion rate and average sequence length on CALVIN D-D from 57.8\% to 67.1\% and from 3.639 to 3.944, respectively. Ablation and real-world experiments support the proposed design, while the real-world evaluation covers a limited number of tasks; future work will explore broader robotic settings.






\section*{ACKNOWLEDGMENT}

ChatGPT (OpenAI) was used during the preparation of this work to assist with language editing and with generating and refining portions of the implementation code, including data processing, and visualization scripts. All AI-assisted code and text were reviewed, tested, and revised by the authors.


\bibliographystyle{IEEEtran}
\bibliography{root}

\end{document}

%% file: figure/overview.tex
\begingroup

\begin{figure*}[t]
    \centering

    \begin{minipage}[c]{0.70\textwidth}
        \centering
        \includegraphics[width=\linewidth]{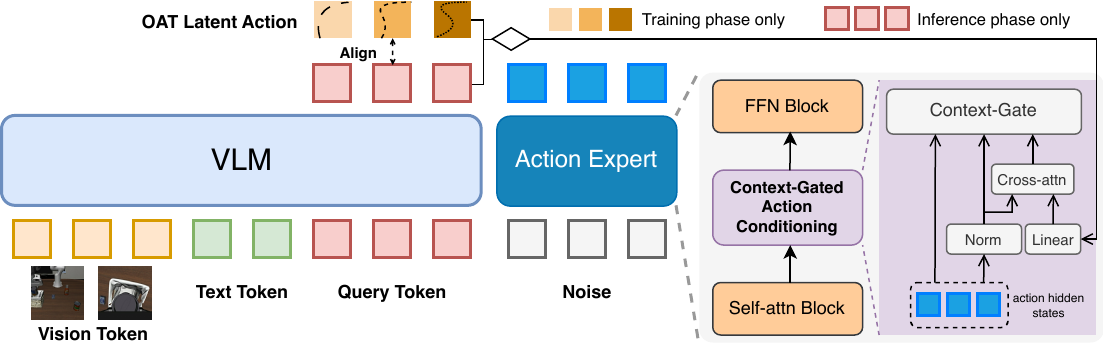}
        \par
        \vspace{-0.3em}
        {\small (a) CAC-VLA Pipeline}
    \end{minipage}
    \hspace{0.02\textwidth}
    \begin{minipage}[c]{0.2\textwidth}
        \centering
        \includegraphics[width=\linewidth]{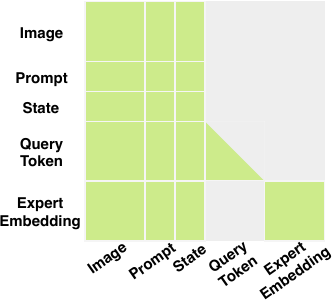}
        \par
        \vspace{-0.3em}
        {\small (b)Attention mask}
    \end{minipage}

    \vspace{-0.3em}
    \caption{
    Overview of CAC-VLA.
    (a) Overall framework of the proposed method.
    (b) Expert-aware attention mask that prevents action expert tokens
    from directly attending to latent-action query tokens.
    }
    \label{fig:overview}
    \vspace{-0.8em}

\end{figure*}

\endgroup

%% file: libero.tex
\begin{table*}[t]
    \centering
    \renewcommand{\arraystretch}{0.95}
    \resizebox{0.88\textwidth}{!}{%
        \begin{tabular}{l c cc cc cc cc cc}
        \toprule
        \multirow{2}{*}{Methods} &
        \multirow{2}{*}{Guidance} &
        \multicolumn{2}{c}{Spatial} &
        \multicolumn{2}{c}{Object} &
        \multicolumn{2}{c}{Goal} &
        \multicolumn{2}{c}{Long} &
        \multicolumn{2}{c}{Avg.} \\
        \cmidrule(lr){3-4}
        \cmidrule(lr){5-6}
        \cmidrule(lr){7-8}
        \cmidrule(lr){9-10}
        \cmidrule(lr){11-12}
        & &
        SR~$\uparrow$ & Rank~$\downarrow$ &
        SR~$\uparrow$ & Rank~$\downarrow$ &
        SR~$\uparrow$ & Rank~$\downarrow$ &
        SR~$\uparrow$ & Rank~$\downarrow$ &
        SR~$\uparrow$ & Rank~$\downarrow$ \\
        \midrule

        UniVLA~\cite{univla} & Visual &
        95.4 & 9 &
        98.8 & 6 &
        93.6 & 9 &
        94.0 & 7 &
        95.5 & 9 \\

        GE-Act~\cite{liao2025genie} & Visual &
        98.2 & 5 &
        97.6 & 9 &
        95.8 & 8 &
        94.4 & 6 &
        96.5 & 7 \\

        \midrule

        $\pi_{0.5}$~\cite{pi05} & Linguistics &
        98.8 & 2 &
        98.2 & 8 &
        98.0 & 3 &
        92.4 & 8 &
        96.9 & 6 \\

        OpenVLA-OFT~\cite{kim2025fine} & Linguistics &
        97.6 & 7 &
        98.4 & 7 &
        97.9 & 4 &
        94.5 & 5 &
        97.1 & 5 \\

        VLA-Adapter~\cite{wang2026vla} & Linguistics &
        97.8 & 6 &
        99.2 & 3 &
        97.2 & 6 &
        95.0 & 3 &
        97.3 & 4 \\

        \midrule

        LAFM~\cite{lafm} & Action &
        97.2 & 8 &
        99.2 & 3 &
        97.2 & 6 &
        92.3 & 9 &
        96.5 & 7 \\

        Coarse-to-Control~\cite{coarse_to_control} & Action &
        98.8 & 2 &
        100.0 & 1 &
        97.8 & 5 &
        95.0 & 3 &
        97.9 & 3 \\

        ACoT-VLA~\cite{acot_vla} & Action &
        98.6 & 4 &
        99.0 & 5 &
        \textbf{99.4} & \textbf{1} &
        97.0 & 2 &
        98.5 & 2 \\

        \rowcolor{cacrowgray}
        \textbf{\sysname} & \textbf{Action} &
        \textbf{99.6} & \textbf{1} &
        \textbf{99.8} & \textbf{2} &
        99.0 & 2 &
        \textbf{97.2} & \textbf{1} &
        \textbf{98.9} & \textbf{1} \\

        \bottomrule
        \end{tabular}%
    }
    \caption{
    Comparison with representative recent methods on the LIBERO benchmark.
    All metrics are average success rates (\%).
    Rankings are computed among the compared methods, and the best results are highlighted in \textbf{bold}.
    Results for LAFM and Coarse-to-Control are reported under their respective LIBERO evaluation protocols.
    }
    \label{tab:libero}
    \vspace{-3mm}
\end{table*}

%% file: libero_plus.tex
\begin{table*}[!t]
    \centering
    \renewcommand{\arraystretch}{0.88}
    \resizebox{0.9\textwidth}{!}{%
    \begin{tabular}{l|c|ccccccc|c}
        \toprule
        Methods & Guidance &
        Camera & Robot & Language & Light & Background & Noise & Layout & Avg. \\
        
        \midrule
        \multicolumn{10}{c}{\textit{Zero-Shot Transfer}} \\
        \midrule
        
        WorldVLA~\cite{worldvla} & Visual &
        0.1 & 27.9 & 41.6 & 43.7 & 17.1 & 10.9 & 38.0 & 25.0 \\
        
        NORA~\cite{hung2025nora} & Linguistics &
        2.2 & 37.0 & 65.1 & 45.7 & 58.6 & 12.8 & 62.1 & 39.0 \\
        
        UniVLA~\cite{univla} & Linguistics &
        1.8 & 46.2 & 69.6 & 69.0 & 81.0 & 21.2 & 31.9 & 42.9 \\
        
        $\pi_0$-Fast~\cite{pertsch2025fast} & Linguistics &
        65.1 & 21.6 & 61.0 & 73.2 & 73.2 & 74.4 & 68.8 & 61.6 \\
        
        RIPT-VLA~\cite{tan2025interactive} & Linguistics &
        55.2 & 31.2 & 77.6 & 88.4 & 91.6 & 73.5 & 74.2 & 68.4 \\
        
        OpenVLA-OFT~\cite{kim2025fine} & Linguistics &
        56.4 & 31.9 & 79.5 & 88.7 & 93.3 & 75.8 & 74.2 & 69.6 \\
        
        
        $\pi_{0.5}^*$~\cite{pi05} & Linguistics &
        65.5 & 75.3 & 84.8 & 95.9 & 92.5 & 81.4 & 87.7 & 82.2 \\


        \rowcolor[gray]{0.92}
        \textbf{\sysname} & \textbf{Action} &
        \textbf{66.6} &
        \textbf{79.4} &
        \textbf{87.0} &
        \textbf{98.3} &
        \textbf{96.3} &
        \textbf{82.6} &
        \textbf{87.3} &
        \textbf{84.2} \\
        
        \midrule
        \multicolumn{10}{c}{\textit{Supervised Fine-Tuning}} \\
        \midrule
    
        
        $\pi_{0.5}^*$~\cite{pi05} & Linguistics &
        92.6 & 71.7 & 81.3 & 95.9 & 96.4 & 93.0 & 86.0 & 87.5 \\

        ACoT-VLA~\cite{acot_vla} & Action &
        \textbf{96.6} &
        70.4 &
        79.7 &
        95.1 &
        97.1 &
        95.9 &
        85.0 &
        88.0 \\

        \rowcolor[gray]{0.92}
        \textbf{\sysname} & \textbf{Action} &
        94.8 &
        \textbf{72.5} &
        \textbf{86.6} &
        \textbf{98.9} &
        \textbf{98.6} &
        \textbf{98.3} &
        \textbf{87.6} &
        \textbf{90.4} \\
        
        \bottomrule
    \end{tabular}
    }
    
    \caption{
    Comparison on LIBERO-Plus.
    \textit{Zero-Shot Transfer} methods are trained on LIBERO and directly evaluated on LIBERO-Plus,
    while \textit{Supervised Fine-Tuning} methods are trained on the LIBERO-Plus training set.
    (*) denotes results reproduced by us for fair comparison.
    Best results are in \textbf{bold}.
    }
    \label{tab:libero_plus}
\end{table*}

%% file: calvin.tex
\begin{table}[t]
    \centering
    \setlength{\tabcolsep}{4pt}
    \renewcommand{\arraystretch}{0.95}
    \begin{tabular}{lcccccc}
        \toprule
        Method
        & 1
        & 2
        & 3
        & 4
        & 5
        & Avg. Len. $\uparrow$ \\
        \midrule
        $\pi_{0.5}$
        & 89.8
        & 79.7
        & 72.2
        & 64.4
        & 57.8
        & 3.639 \\

        \rowcolor{cacrowgray}
        \textbf{\sysname}
        & \textbf{91.5}
        & \textbf{84.6}
        & \textbf{78.3}
        & \textbf{72.9}
        & \textbf{67.1}
        & \textbf{3.944} \\
        \bottomrule
    \end{tabular}
    \caption{Comparison on the CALVIN D-D benchmark. Columns 1--5 report the success rates for completing 1--5 consecutive tasks, respectively. Avg. Len. denotes the average sequence length. The $\pi_{0.5}$ result is reproduced by us for fair comparison.}
    \label{tab:calvin}
    \vspace{-5.0mm}
\end{table}

%% file: abl_horizon.tex
\begin{table}[t]
    \centering
    \scriptsize
    \setlength{\tabcolsep}{2pt}
    \renewcommand{\arraystretch}{1.2}
    \resizebox{\columnwidth}{!}{%
    \begin{tabular}{l|ccccccc|c}
        \toprule
        $H_l$ & Camera & Robot & Language & Light & Bkg. & Noise & Layout & Avg. \\
        \midrule
        10 & \textbf{94.81} & \textbf{72.45} & \textbf{86.60} & \textbf{98.86} & \textbf{98.61} & \textbf{98.31} & \textbf{87.61} & \textbf{90.43} \\
        20 & 94.56 & 70.52 & 81.91 & 97.81 & 95.26 & 96.25 & 86.43 & 88.38 \\
        30 & 94.37 & 68.06 & 79.83 & 97.20 & 97.30 & 94.50 & 87.28 & 87.66 \\
        \bottomrule
    \end{tabular}%
    }
    \caption{Ablation study on the latent-action horizon on LIBERO-Plus. All metrics are success rates (\%).}
    \label{tab:ablation_libero_plus_horizon}
    \vspace{-4.0mm}
\end{table}

%% file: abl_module.tex
\begin{table}[t]
    \centering
    \scriptsize
    \setlength{\tabcolsep}{2pt}
    \renewcommand{\arraystretch}{1.2}
    \resizebox{\columnwidth}{!}{%
    \begin{tabular}{l|ccccccc|c}
        \toprule
        Variant & Camera & Robot & Language & Light & Bkg. & Noise & Layout & Avg. \\
        \midrule
        \sysname & 94.81 & \textbf{72.45} & \textbf{86.60} & 98.86 & \textbf{98.61} & \textbf{98.31} & \textbf{87.61} & \textbf{90.43} \\
        w/o Context-Gate & \textbf{94.87} & 71.55 & 81.91 & \textbf{99.21} & 98.51 & 97.50 & 86.82 & 89.36 \\
        w/o Latent & 94.25 & 70.90 & 82.76 & 95.53 & 95.72 & 94.19 & 86.16 & 87.95 \\
        \bottomrule
    \end{tabular}%
    }
    \caption{Ablation study of the context-gated action conditioning module on LIBERO-Plus. ``w/o Context-Gate'' removes the context gate, while ``w/o Latent'' removes latent-action conditioning. All metrics are success rates (\%).}
    \label{tab:ablation_libero_plus_module}
    \vspace{-4.0mm}
\end{table}

%% file: figure/real_merge.tex
\begin{figure*}[t]
\centering

\begin{minipage}[c]{0.55\textwidth}
    \centering

    \includegraphics[width=\linewidth]{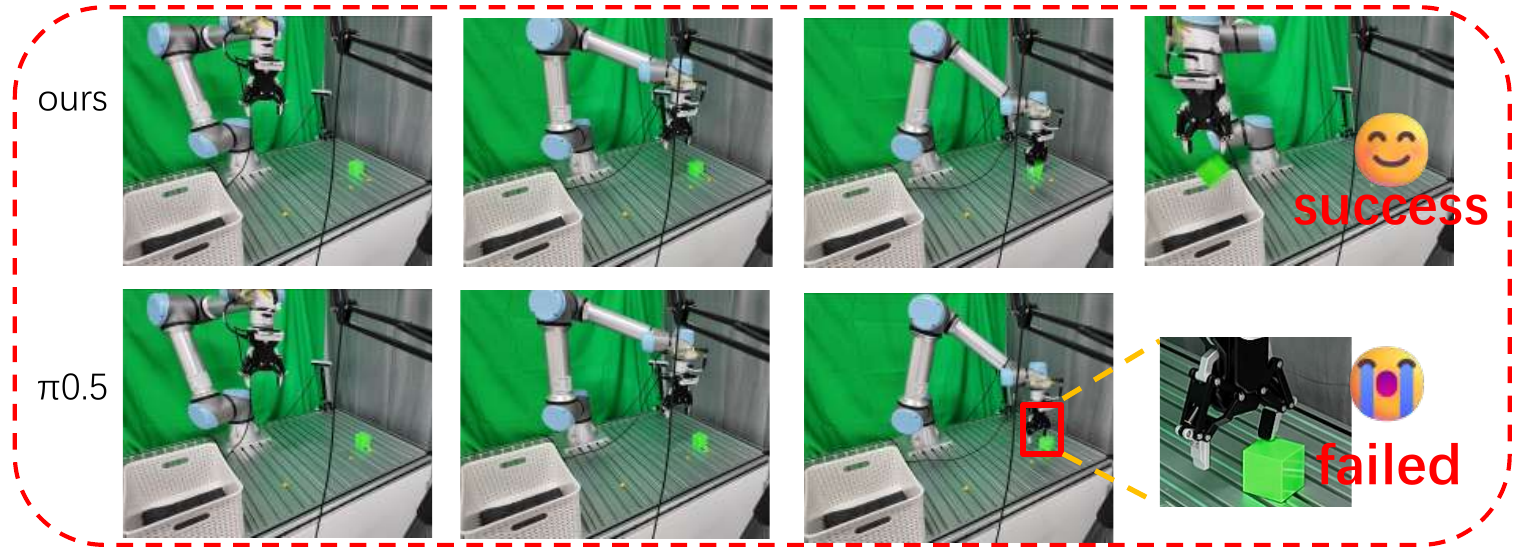}
    \par
    \vspace{-0.3em}
    {\fontsize{7.2}{8.2}\selectfont (a) Pick-and-place setup}

    \vspace{0.5em}

    \includegraphics[width=\linewidth]{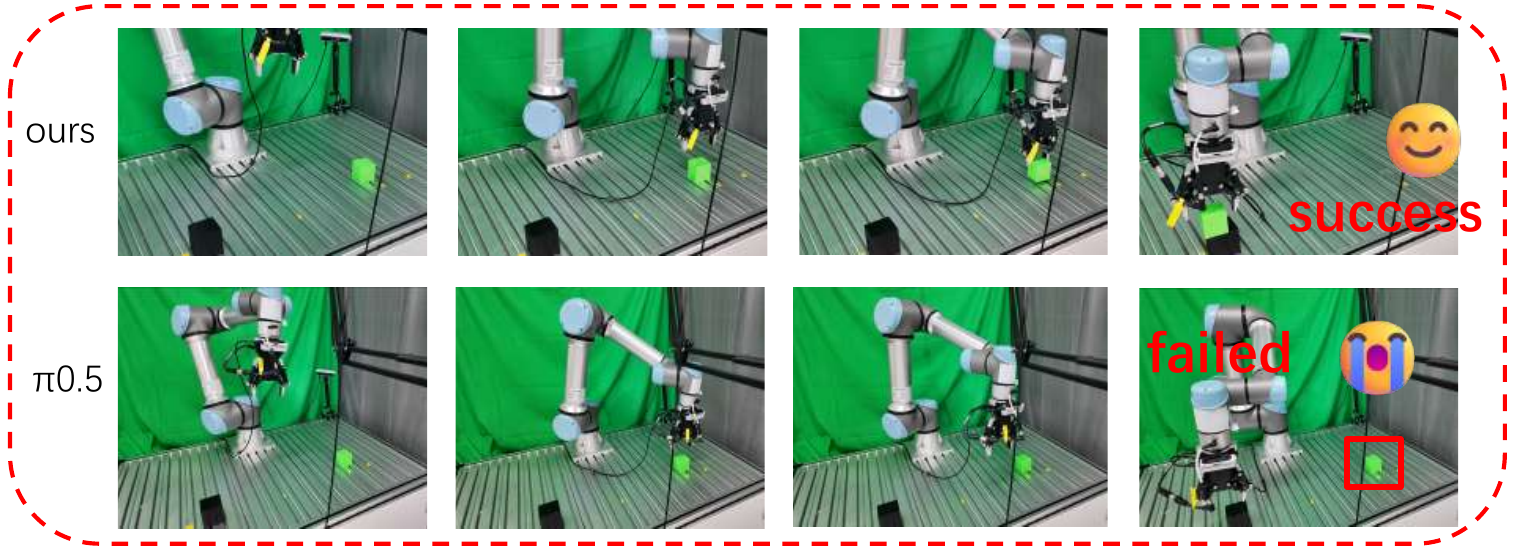}
    \par
    \vspace{-0.3em}
    {\fontsize{7.2}{8.2}\selectfont (b) Block-stacking setup}
\end{minipage}
\hspace{0.025\textwidth}
%
\begin{minipage}[c]{0.4\textwidth}
    \centering
    \includegraphics[width=\linewidth]{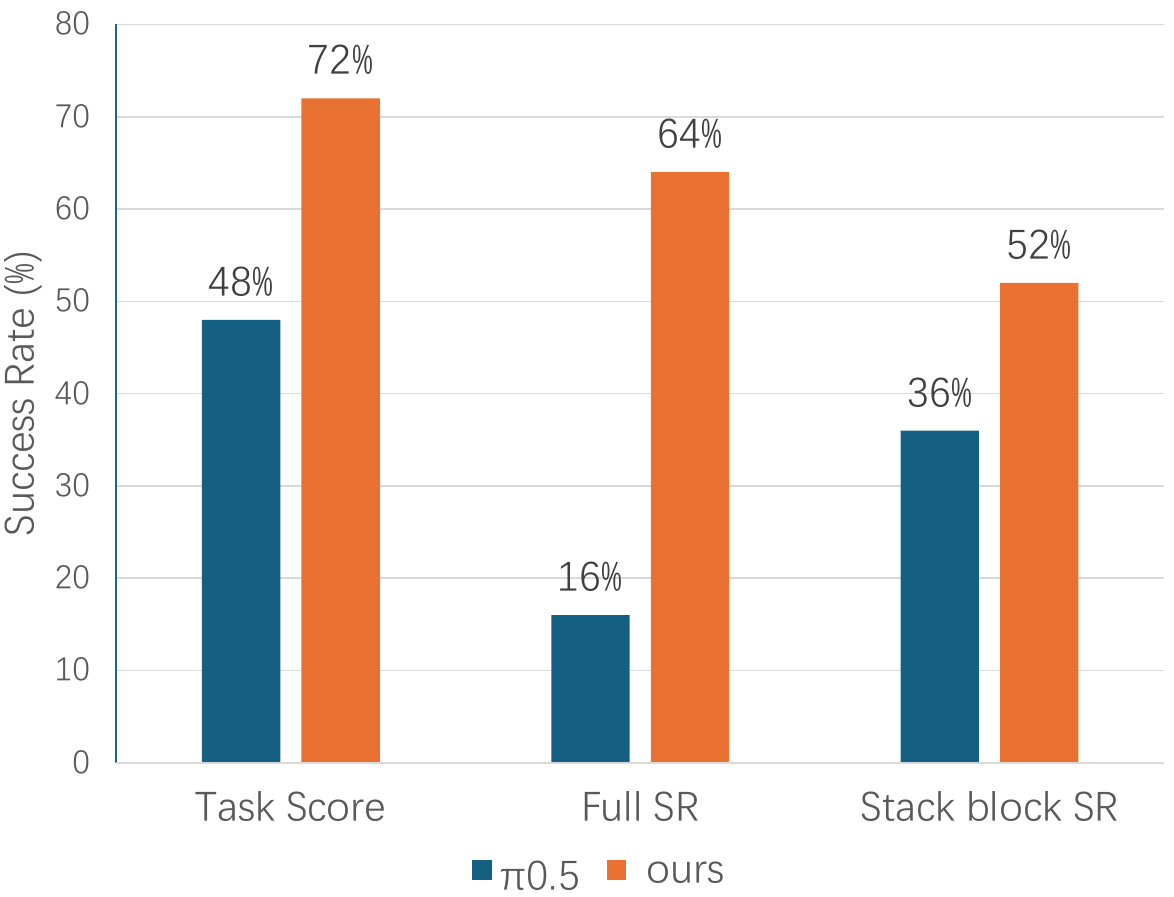}
    \par
    \vspace{-0.3em}
    {\fontsize{7.2}{8.2}\selectfont (c) Evaluation results}
\end{minipage}

\vspace{-0.3em}
\caption{Quantitative real-world evaluation on pick-and-place and block-stacking tasks.
We report Task Score and Full Success Rate (Full SR) for pick-and-place and Stacking Success Rate (Stacking SR) for block stacking.}
\label{fig:real_results}
\vspace{-1.0em}

\end{figure*}

%% file: root.bib
@misc{paligemma,
      title={PaliGemma: A versatile 3B VLM for transfer}, 
      author={Lucas Beyer and Andreas Steiner and André Susano Pinto and Alexander Kolesnikov and Xiao Wang and Daniel Salz and Maxim Neumann and Ibrahim Alabdulmohsin and Michael Tschannen and Emanuele Bugliarello and Thomas Unterthiner and Daniel Keysers and Skanda Koppula and Fangyu Liu and Adam Grycner and Alexey Gritsenko and Neil Houlsby and Manoj Kumar and Keran Rong and Julian Eisenschlos and Rishabh Kabra and Matthias Bauer and Matko Bošnjak and Xi Chen and Matthias Minderer and Paul Voigtlaender and Ioana Bica and Ivana Balazevic and Joan Puigcerver and Pinelopi Papalampidi and Olivier Henaff and Xi Xiong and Radu Soricut and Jeremiah Harmsen and Xiaohua Zhai},
      year={2024},
      eprint={2407.07726},
      archivePrefix={arXiv},
      primaryClass={cs.CV},
      url={https://arxiv.org/abs/2407.07726}, 
}

@misc{swiftVLA,
      title={SwiftVLA: Unlocking Spatiotemporal Dynamics for Lightweight VLA Models at Minimal Overhead}, 
      author={Chaojun Ni and Cheng Chen and Xiaofeng Wang and Zheng Zhu and Wenzhao Zheng and Boyuan Wang and Tianrun Chen and Guosheng Zhao and Haoyun Li and Zhehao Dong and Qiang Zhang and Yun Ye and Yang Wang and Guan Huang and Wenjun Mei},
      year={2025},
      eprint={2512.00903},
      archivePrefix={arXiv},
      primaryClass={cs.CV},
      url={https://arxiv.org/abs/2512.00903}, 
}

@misc{oat,
      title={OAT: Ordered Action Tokenization}, 
      author={Chaoqi Liu and Xiaoshen Han and Jiawei Gao and Yue Zhao and Haonan Chen and Yilun Du},
      year={2026},
      eprint={2602.04215},
      archivePrefix={arXiv},
      primaryClass={cs.RO},
      url={https://arxiv.org/abs/2602.04215}, 
}

@article{acot_vla,
  title={ACoT-VLA: Action Chain-of-Thought for Vision-Language-Action Models},
  author={Zhong, Linqing and Liu, Yi and Wei, Yifei and Xiong, Ziyu and Yao, Maoqing and Liu, Si and Ren, Guanghui},
  journal={arXiv preprint arXiv:2601.11404},
  year={2026}
}

@article{palm_e,
  title={PALM-E: An Embodied Multimodal Language Model},
  author={Driess, Danny and Xia, Fei and Sajjadi, Mehdi S. M. and Lynch, Corey and Chowdhery, Aakanksha and Ichter, Brian and Wahid, Ayzaan and Tompson, Jonathan and Vuong, Quan and Yu, Tianhe and Huang, Wenlong and Chebotar, Yevgen and Sermanet, Pierre and Duckworth, Daniel and Levine, Sergey and Vanhoucke, Vincent and Hausman, Karol and Toussaint, Marc and Greff, Klaus and Zeng, Andy and Mordatch, Igor and Florence, Pete},
  journal={arXiv preprint arXiv:2303.03378},
  year={2023},
  url={https://doi.org/10.48550/arXiv.2303.03378}
}

@article{openvla,
    title={OpenVLA: An Open-Source Vision-Language-Action Model},
    author={{Moo Jin} Kim and Karl Pertsch and Siddharth Karamcheti and Ted Xiao and Ashwin Balakrishna and Suraj Nair and Rafael Rafailov and Ethan Foster and Grace Lam and Pannag Sanketi and Quan Vuong and Thomas Kollar and Benjamin Burchfiel and Russ Tedrake and Dorsa Sadigh and Sergey Levine and Percy Liang and Chelsea Finn},
    journal = {arXiv preprint arXiv:2406.09246},
    year={2024}
}

@inproceedings{octo,
    title={Octo: An Open-Source Generalist Robot Policy},
    author = {{Octo Model Team} and Dibya Ghosh and Homer Walke and Karl Pertsch and Kevin Black and Oier Mees and Sudeep Dasari and Joey Hejna and Charles Xu and Jianlan Luo and Tobias Kreiman and {You Liang} Tan and Pannag Sanketi and Quan Vuong and Ted Xiao and Dorsa Sadigh and Chelsea Finn and Sergey Levine},
    booktitle = {Proceedings of Robotics: Science and Systems},
    address  = {Delft, Netherlands},
    year = {2024},
}

@misc{pi0,
      title={$\pi_0$: A Vision-Language-Action Flow Model for General Robot Control}, 
      author={Kevin Black and Noah Brown and Danny Driess and Adnan Esmail and Michael Equi and Chelsea Finn and Niccolo Fusai and Lachy Groom and Karol Hausman and Brian Ichter and Szymon Jakubczak and Tim Jones and Liyiming Ke and Sergey Levine and Adrian Li-Bell and Mohith Mothukuri and Suraj Nair and Karl Pertsch and Lucy Xiaoyang Shi and James Tanner and Quan Vuong and Anna Walling and Haohuan Wang and Ury Zhilinsky},
      year={2026},
      eprint={2410.24164},
      archivePrefix={arXiv},
      primaryClass={cs.LG},
      url={https://arxiv.org/abs/2410.24164}, 
}

@misc{pi05,
      title={$\pi_{0.5}$: a Vision-Language-Action Model with Open-World Generalization}, 
      author={Physical Intelligence and Kevin Black and Noah Brown and James Darpinian and Karan Dhabalia and Danny Driess and Adnan Esmail and Michael Equi and Chelsea Finn and Niccolo Fusai and Manuel Y. Galliker and Dibya Ghosh and Lachy Groom and Karol Hausman and Brian Ichter and Szymon Jakubczak and Tim Jones and Liyiming Ke and Devin LeBlanc and Sergey Levine and Adrian Li-Bell and Mohith Mothukuri and Suraj Nair and Karl Pertsch and Allen Z. Ren and Lucy Xiaoyang Shi and Laura Smith and Jost Tobias Springenberg and Kyle Stachowicz and James Tanner and Quan Vuong and Homer Walke and Anna Walling and Haohuan Wang and Lili Yu and Ury Zhilinsky},
      year={2025},
      eprint={2504.16054},
      archivePrefix={arXiv},
      primaryClass={cs.LG},
      url={https://arxiv.org/abs/2504.16054}, 
}

@article{saycan,
  title={Do as i can, not as i say: Grounding language in robotic affordances},
  author={Ahn, Michael and Brohan, Anthony and Brown, Noah and Chebotar, Yevgen and Cortes, Omar and David, Byron and Finn, Chelsea and Fu, Chuyuan and Gopalakrishnan, Keerthana and Hausman, Karol and others},
  journal={arXiv preprint arXiv:2204.01691},
  year={2022}
}

@misc{cot_vla,
      title={CoT-VLA: Visual Chain-of-Thought Reasoning for Vision-Language-Action Models}, 
      author={Qingqing Zhao and Yao Lu and Moo Jin Kim and Zipeng Fu and Zhuoyang Zhang and Yecheng Wu and Zhaoshuo Li and Qianli Ma and Song Han and Chelsea Finn and Ankur Handa and Ming-Yu Liu and Donglai Xiang and Gordon Wetzstein and Tsung-Yi Lin},
      year={2025},
      eprint={2503.22020},
      archivePrefix={arXiv},
      primaryClass={cs.CV},
      url={https://arxiv.org/abs/2503.22020}, 
}

@article{worldvla,
  title={WorldVLA: Towards Autoregressive Action World Model},
  author={Cen, Jun and Yu, Chaohui and Yuan, Hangjie and Jiang, Yuming and Huang, Siteng and Guo, Jiayan and Li, Xin and Song, Yibing and Luo, Hao and Wang, Fan and others},
  journal={arXiv preprint arXiv:2506.21539},
  year={2025}
}

@misc{dreamvla,
      title={DreamVLA: A Vision-Language-Action Model Dreamed with Comprehensive World Knowledge}, 
      author={Wenyao Zhang and Hongsi Liu and Zekun Qi and Yunnan Wang and Xinqiang Yu and Jiazhao Zhang and Runpei Dong and Jiawei He and Fan Lu and He Wang and Zhizheng Zhang and Li Yi and Wenjun Zeng and Xin Jin},
      year={2025},
      eprint={2507.04447},
      archivePrefix={arXiv},
      primaryClass={cs.CV},
      url={https://arxiv.org/abs/2507.04447}, 
}

@article{lapa,
  title={Latent Action Pretraining from Videos},
  author={Ye, Seonghyeon and Jang, Joel and Jeon, Byeongguk and Joo, Sejune and Yang, Jianwei and Peng, Baolin and Mandlekar, Ajay and Tan, Reuben and Chao, Yu-Wei and Lin, Bill Yuchen and others},
  journal={arXiv preprint arXiv:2410.11758},
  year={2024}
}

@article{univla,
  title={Univla: Learning to act anywhere with task-centric latent actions},
  author={Bu, Qingwen and Yang, Yanting and Cai, Jisong and Gao, Shenyuan and Ren, Guanghui and Yao, Maoqing and Luo, Ping and Li, Hongyang},
  journal={arXiv preprint arXiv:2505.06111},
  year={2025}
}

@article{liao2025genie,
  title={Genie envisioner: A unified world foundation platform for robotic manipulation},
  author={Liao, Yue and Zhou, Pengfei and Huang, Siyuan and Yang, Donglin and Chen, Shengcong and Jiang, Yuxin and Hu, Yue and Cai, Jingbin and Liu, Si and Luo, Jianlan and others},
  journal={arXiv preprint arXiv:2508.05635},
  year={2025}
}

@article{pertsch2025fast,
  title={Fast: Efficient action tokenization for vision-language-action models},
  author={Pertsch, Karl and Stachowicz, Kyle and Ichter, Brian and Driess, Danny and Nair, Suraj and Vuong, Quan and Mees, Oier and Finn, Chelsea and Levine, Sergey},
  journal={arXiv preprint arXiv:2501.09747},
  year={2025}
}

@article{kim2025fine,
  title={Fine-tuning vision-language-action models: Optimizing speed and success},
  author={Kim, Moo Jin and Finn, Chelsea and Liang, Percy},
  journal={arXiv preprint arXiv:2502.19645},
  year={2025}
}

@inproceedings{wang2026vla,
  title={Vla-adapter: An effective paradigm for tiny-scale vision-language-action model},
  author={Wang, Yihao and Ding, Pengxiang and Li, Lingxiao and Cui, Can and Ge, Zirui and Tong, Xinyang and Song, Wenxuan and Zhao, Han and Zhao, Wei and Hou, Pengxu and others},
  booktitle={Proceedings of the AAAI conference on artificial intelligence},
  volume={40},
  pages={18638--18646},
  year={2026}
}

@article{hung2025nora,
  title={Nora: A small open-sourced generalist vision language action model for embodied tasks},
  author={Hung, Chia-Yu and Sun, Qi and Hong, Pengfei and Zadeh, Amir and Li, Chuan and Tan, U and Majumder, Navonil and Poria, Soujanya and others},
  journal={arXiv preprint arXiv:2504.19854},
  year={2025}
}

@article{tan2025interactive,
  title={Interactive Post-Training for Vision-Language-Action Models},
  author={Tan, Shuhan and Dou, Kairan and Zhao, Yue and Kr{\"a}henb{\"u}hl, Philipp},
  journal={arXiv preprint arXiv:2505.17016},
  year={2025}
}

@article{lafm,
  title={Flowing With Purpose: Latent Action Guided Flow Matching Policies for Robotic Manipulation},
  author={Machado, Bruno and Chapin, Alexandre and Dellandrea, Emmanuel and Chen, Liming},
  journal={arXiv preprint arXiv:2606.23420},
  year={2026}
}

@article{coarse_to_control,
  title={Coarse-to-Control: Action-Token Planning for Vision-Language-Action Models},
  author={Wu, Jinhao and Zhang, Shiduo and Liu, Yicheng and Yu, Xiaopeng and Li, Sixian and Wang, Siyin and Zhao, Hang and Huo, Jing and Gao, Yang and Gong, Jingjing and Qiu, Xipeng and Jiang, Yu-Gang},
  journal={arXiv preprint arXiv:2606.07107},
  year={2026}
}

@article{libero,
  title={Libero: Benchmarking knowledge transfer for lifelong robot learning},
  author={Liu, Bo and Zhu, Yifeng and Gao, Chongkai and Feng, Yihao and Liu, Qiang and Zhu, Yuke and Stone, Peter},
  journal={Advances in Neural Information Processing Systems},
  volume={36},
  pages={44776--44791},
  year={2023}
}

@article{libero_plus,
  title={LIBERO-Plus: In-depth Robustness Analysis of Vision-Language-Action Models},
  author={Fei, Senyu and Wang, Siyin and Shi, Junhao and Dai, Zihao and Cai, Jikun and Qian, Pengfang and Ji, Li and He, Xinzhe and Zhang, Shiduo and Fei, Zhaoye and others},
  journal={arXiv preprint arXiv:2510.13626},
  year={2025}
}

@article{calvin,
author = {Oier Mees and Lukas Hermann and Erick Rosete-Beas and Wolfram Burgard},
title = {CALVIN: A Benchmark for Language-Conditioned Policy Learning for Long-Horizon Robot Manipulation Tasks},
journal={IEEE Robotics and Automation Letters (RA-L)},
volume={7},
number={3},
pages={7327-7334},
year={2022}
}

@inproceedings{yu2024ldp,
  title={Ldp: A local diffusion planner for efficient robot navigation and collision avoidance},
  author={Yu, Wenhao and Peng, Jie and Yang, Huanyu and Zhang, Junrui and Duan, Yifan and Ji, Jianmin and Zhang, Yanyong},
  booktitle={2024 IEEE/RSJ International Conference on Intelligent Robots and Systems (IROS)},
  pages={5466--5472},
  year={2024},
  organization={IEEE}
}

@article{duan2025stdarm,
  title={STDArm: Transferring visuomotor policies from static data training to dynamic robot manipulation},
  author={Duan, Yifan and Li, Heng and Wu, Yilong and Yu, Wenhao and Zhang, Xinran and Shen, Yedong and Ji, Jianmin and Zhang, Yanyong},
  journal={arXiv preprint arXiv:2504.18792},
  year={2025}
}

@article{gao2026drift,
  title={Drift-Based Policy Optimization: Native One-Step Policy Learning for Online Robot Control},
  author={Gao, Yuxuan and Shen, Yedong and Zhang, Shiqi and Yu, Wenhao and Duan, Yifan and Wu, Jiajia and Deng, Jiajun and Zhang, Yanyong and others},
  journal={arXiv preprint arXiv:2604.03540},
  year={2026}
}
